\documentclass{article}
\usepackage{ijcai22}

\usepackage{times}

\usepackage{soul}
\usepackage{url}
\usepackage[hidelinks]{hyperref}
\usepackage[utf8]{inputenc}
\usepackage[small]{caption}
\usepackage{graphicx}
\usepackage{amsmath}
\usepackage{xcolor}
\usepackage{amsthm}
\usepackage{algorithm}
\usepackage{subcaption}
\usepackage{booktabs}
\usepackage{multicol}
\usepackage{multirow}
\usepackage{colortbl}
\usepackage[noend]{algpseudocode}
\newtheorem*{remark*}{Assertion}

\usepackage{booktabs}
\title{Process Knowledge-infused Learning for Suicidality Assessment on Social Media}
\author{
Kaushik Roy$^{1}$\footnote{Contact Author}\and
Manas Gaur$^{1}$\and
Qi Zhang$^{1}$\And
Amit Sheth$^{1}$\\
\affiliations
$^1$University of South Carolina, Artificial Intelligence Institute, SC, USA
\emails
\{kaushikr, mgaur\}@email.sc.edu,
qz5@cse.sc.edu,
amit@sc.edu
}

\begin{document}

\maketitle

\begin{abstract}
Improving the performance and natural language explanations of deep learning algorithms is a priority for adoption by humans in the real world. In several domains, such as healthcare, such technology has significant potential to reduce the burden on humans by providing quality assistance at scale. However, current methods rely on the traditional pipeline of predicting labels from data, thus completely ignoring the process and guidelines used to obtain the labels. Furthermore, post hoc explanations on the data to label prediction using explainable AI (XAI) models, while satisfactory to computer scientists, leave much to be desired to the end-users due to lacking explanations of the process in terms of human-understandable concepts. We \textit{introduce}, \textit{formalize}, and \textit{develop} a novel Artificial Intelligence (A) paradigm - Process Knowledge-infused Learning (PK-iL). PK-iL utilizes a structured process knowledge that explicitly explains the underlying prediction process that makes sense to end-users. The qualitative human evaluation con-
firms through a annotator agreement of 0.72, that humans are understand explanations for the predictions. PK-iL also performs competitively with the state-of-the-art (SOTA) baselines. 
\end{abstract}


\section{Introduction}
\begin{figure*}[ht]
    \centering
\begin{subfigure}[b]{.67\textwidth}
  \centering
  \includegraphics[width=\linewidth]{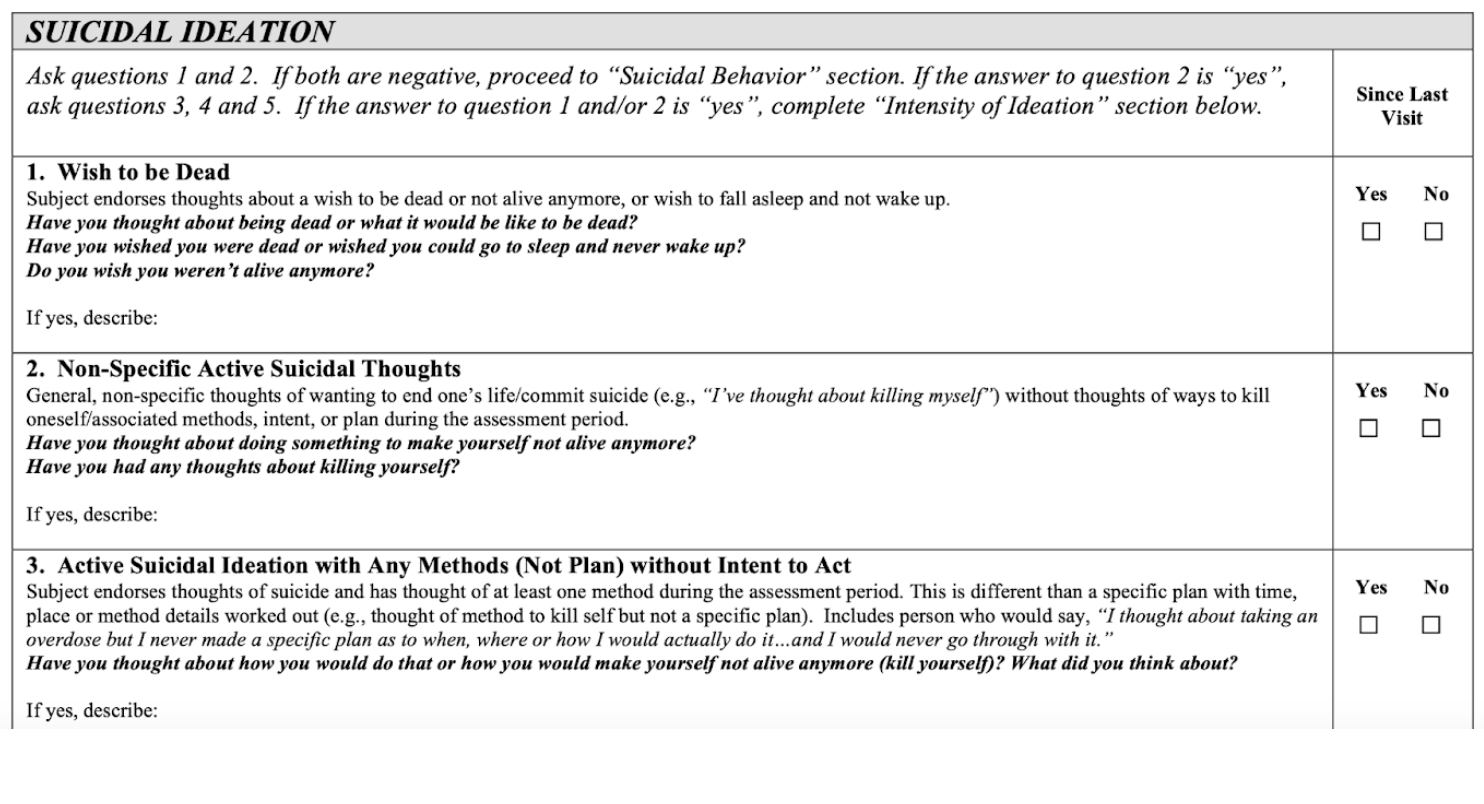}
\end{subfigure}
\hfill
\begin{subfigure}[b]{.31\textwidth}
  \centering
  \includegraphics[width=\linewidth]{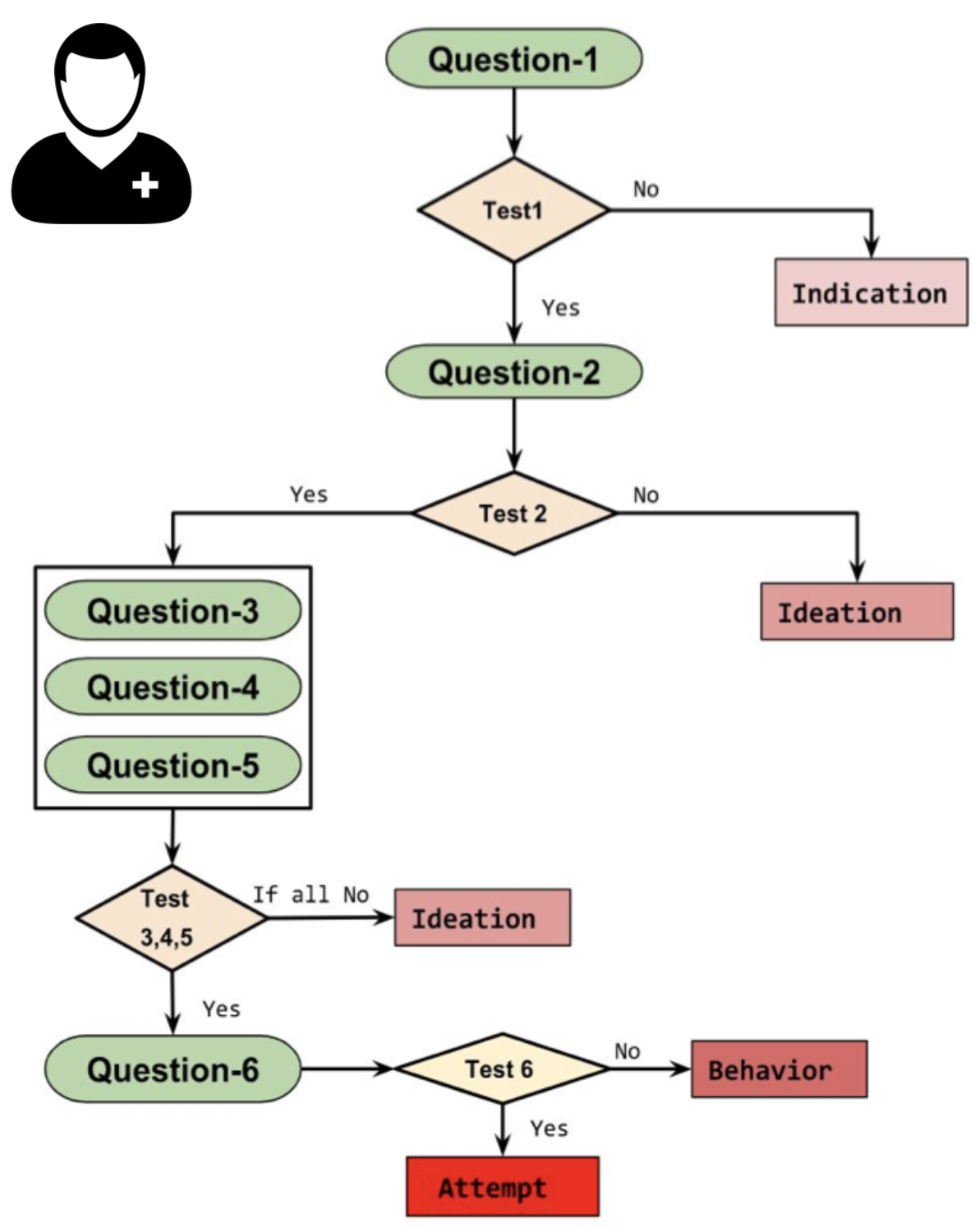}
\end{subfigure}
\caption{\textit{Left:} Columbia Suicide Severity Rating Scale. \textit{Right:} The induced process knowledge.}
\label{fig:cssrs}
\end{figure*}

{\label{sec:intro}}
A long-standing problem in adopting machine learning technologies to assist humans in the real world has been the lack of a satisfactory explanation to the end-users of the technology. In the traditional machine learning pipeline, much attention is paid to fitting a function map from data points to labels. However, during the annotation of data points in the ground truth dataset, a guideline or process is often detailed by which the annotator can label the dataset. For example, to label patients for degrees of suicidal tendencies in a physical clinical setting, a well-known scale, the Columbia Suicide Severity Rating Scale (CSSRS) \cite{bjureberg2021columbia}, is used to determine the right set of labels. Figure \ref{fig:cssrs} shows this scale. Thus, it is clear to the patient how a particular suicidal tendency is recognized once the clinician evaluates the questions and patient responses.

Similarly, when data points in a dataset are annotated in other domains, each data point is evaluated against a process or guideline similar to the CSSRS by several human annotators. The assumption is that the machine learning algorithm will implicitly recover the underlying process or guideline used by the annotators when learning a function map from data point to label. Popular methods for XAI such as LIME and SHaP, are used to explain the function learned, often through local approximations related to a single or sampled set of data points \cite{adadi2018peeking,ribeiro2016should,lundberg2017unified}. However, due to the black-box nature of the function and the non-convexity of the hypothesis function surfaces, it is challenging to evaluate if the recovery of the underlying process or guideline was successful and is meaningful to the end-users. Fundamentally, we might think of these XAI methods as trying to roughly construct an explanation as saying, ``This data point is explainable using a simpler hypothesis function (a local approximation) due to \textit{similar} data points (data points in the local neighborhood) also being classified correctly by the simpler hypothesis''. Consequently, much depends on the choice of local approximation and the machine learning models understanding of \textit{similar} data points, on what is already a highly non-convex gargantuan function such as a large SOTA language model (LM) \cite{vaswani2017attention}. Also, while such an explanation may satisfy the computer science community, ``similarities'' are hardly adequate for the end-user (e.g., a psychotherapist). The pertinent questions include: Would the human annotators consider the data points deemed \textit{similar} by the LM also to be similar to each other? Would the human annotators agree that the explanation by the local approximation is aligned with the process or guideline used by them to label those data points?

In our study, we ask the question, what if we were able to use not just the annotator's labels, but also the process or guidelines used to label them and explicitly control the learning of a model to recover the process or guideline (instead of implicitly). Such an algorithm would, by design, be explainable and emulate the humans model of \textit{similarity} between data points. 
This paper takes the first step to answer this question grounded in the deep learning task of suicidality assessment from social media data, where incorporating the knowledge of medical processes and guidelines is of critical considerations.
To this end, we propose a novel class of algorithms \textbf{Process Knowledge infused Learning (PK-iL)} for suicidality assessment from social media. We make the following contributions:
\begin{itemize}
    \item Define \textbf{Process Knowledge (PK)} and create a dataset for suicidality assessment task based on CSSRS with annotations to include both PK and labels.
    \item Develop \textbf{Process Knowledge infused Learning (PK-iL)}, an explainable algorithm that explicitly controls the learning model to recover the process by effective utilization of PK in the annotation and a globally optimal optimization objective.
\end{itemize}

\paragraph{Can PK-iL utilize SOTA LMs?}
We note that the notion of \textit{similar} data points is the machine learning model's way of understanding the human annotator's annotation process, i.e., fundamentally, the goal of the model is constructing a similarity space that mimics the human annotator's understanding. In many domains and applications, large and SOTA LMs have excelled at capturing the similarity of some examples exceptionally well. Hence, we believe that rather than try to implicitly learn a \textit{similarity} space as a model of the human annotator's understanding over the whole space of examples, we can leverage SOTA LMs to define the annotator's understanding at process-specific checkpoints in the \textbf{PK}. For instance, if a \textbf{PK} has five questions (or guidelines) to go through, we can use the SOTA model as a proxy to understand if a human annotator would have judged the guideline as satisfied. Such finer-grained understanding can potentially leverage the ability of SOTA models to learn similarity spaces while still maintaining the explicitly explainable \textbf{PK-iL} structure. 

\paragraph{}
We will see the data collection, examples and intuitions, and the formalization of \textbf{PK-iL} in action through the following sections.

\section{PK enhanced CSSRS dataset}{\label{sec:dataset}}
Before the pandemic, suicidality was already a  leading mental health issue across the world. Since the pandemic, incidents of suicidality have increased even further. Thus for both demonstrating high real-world impact through an important use case on real data and users as well as for ease of exposition, we explain our methods and experiments anchored around the application of suicidal thought pattern detection. However, \textbf{PK-iL} is generalizable to any domain that requires the integration of \textbf{PK} with data to derive high-quality explanations. 

To conduct our study in a physical, real-world experimentation setting, we require responses from users physically present during the experiment. Consider the clinical setting of suicidal thought assessments using the CSSRS - obtaining access to a physical clinical setting presents many hurdles such as ethics approval, incentives for honest responses, etc. The demand-supply deficit in mental health already makes it hard to find a quality experimental setting, and the recent COVID-19 pandemic has compounded this issue. The significant number of persons turning to social media platforms presents an exciting opportunity to leverage a large amount of data as a proxy for user responses. Thus we utilize the dataset of Gaur et al., \cite{gaur2021characterization,alambo2019question} which uses the CSSRS to label user posts from suicide-related subreddits and thus provides a real-world test-bed to evaluate the performance and explainability of \textbf{PK-iL}. Through the CSSRS, the domain experts in the study annotated longitudinal data from 448 users for the following labels: Suicide Ideation, Suicide Behavior, Suicide Attempt. High standards in annotation were maintained with a substantial inter-rater agreement of 0.84. Crucially, we expand this dataset to include the specific guideline (\textbf{PK}) for annotation in addition to the label. Table \ref{table:examples} shows examples of the dataset expanded with \textbf{PK}.

\begin{table*}[!htbp]
\footnotesize
\centering
\begin{tabular}{p{6cm}|p{6cm}|p{4cm}} 
\toprule[1.5pt]
User text (x)                                                                                      & \color{gray}Process Knowledge (\textbf{PK})
& \begin{tabular}[c]{@{}l@{}}CSSRS Label\end{tabular}  \\ 
\midrule[1pt]
\multirow{3}{*}{\begin{tabular}[c]{@{}l@{}} [...] a voice telling me to kill myself [...] \\  yes - I think I should do it [...] \end{tabular}}       & 1.2 (yes - a voice telling me to kill myself) \\ & 2.2 (yes - I think I should do it) \\ & 4 (yes - I think I should do it)                          & Behavior or Attempt                                                                                            \\ 
\hline
\multirow{2}{*}{\begin{tabular}[c]{@{}l@{}}[...] Rarely is a day \\ where I dont suffer from thoughts of self-harm...
\end{tabular}} & 1.1 (yes - Rarely is a day where I dont suffer from thoughts of self-harm)  \\ & 2 (no - no words indicating active suicidal thought)                 & Ideation                                          \\
\bottomrule[1.5pt]
\end{tabular}
\caption{Examples of data set annotation expanded with \textbf{PK}. The [...] collapses the rest of the post for brevity. Each question (1-6) in the CSSRS has a main question and sub questions 1.1, 1.2 etc, as can be seen in Figure \ref{fig:cssrs}. Thus the \textbf{PK} denotes the main question or sub-question being answered in the user's Reddit post.}
\label{table:examples}
\end{table*}

\section{Tree Structure of PK}{\label{sec:treestruct}}
We see in Figure \ref{fig:cssrs} that \textbf{PK} can be viewed as a decision tree. A process tree (\textbf{Process Knowledge}(PK)) to to determine the probability of a label $y$ for a user post can be written as a polynomial of the form 
\begin{equation}
y = \sum_{l\in Leaves}p_{l}\prod_{i=1}^{N_{q}}\big(I_{yes}(q_{i}))\big(1-I_{no}(q_{i}))
\label{eqn:treeprob}
\end{equation}, where $N_{q
}$ is the number of questions in the decision tree, $I_{yes}(q_{i})$ and $I_{no}(q_{i})$ represents if the post follows a \textit{yes} path or a \textit{no} path to the question $q_{i}$. $Leaves$ is a set of all leaves that lead to the label $y$. For example, there are two paths in Figure \ref{fig:cssrs} that lead to $y=Ideation$. Here $p_{l}$ is computed as the ratio of the number of annotators that chose that path for the example to the total number of annotators - this in some sense captures the inter annotator agreement for those set of examples. For example, considering a particular post, if among three annotators, two annotators labeled the \textbf{PK} as the path $1.2\rightarrow 2.2\rightarrow 4$. Then the probability of $y = Behavior or Attempt$ for that post is $0.66$. Note here that the sub questions aren't stored in the tree leaves. The path $1.2\rightarrow 2.2\rightarrow 4$ is equivalent to $1\rightarrow 2 \rightarrow4$. This is done for all the examples in the training set and the final probability is an average of all the examples.
\begin{remark*}
For any model $\mathcal{M}(y)$ that approximates the probability of label $y$ for a post according to Equation \ref{eqn:treeprob}, let the inter-annotator for the post labeled as $y$ be $\mathcal{A}(y)$. Then best approximation for the post, $\mathcal{M}^{*}(y) \leq \mathcal{A}(y)$
\label{rem:first}
\end{remark*}
We claim the above as an assertion instead of a theorem as it is trivial to see that $\prod_{i=1}^{N_{q}}\big(I_{yes}(q_{i}))\big(1-I_{no}(q_{i})) \leq 1$ always, and therefore any approximation is upper bounded by the inter-annotator agreement. This makes intuitive sense as what we are really interested in capturing the annotator's thought process while labeling and on par accuracy. Improving upon the inter-annotator agreement may mean capturing something that is not present in the ground truth. Thus, we are interested in labeling unseen data as well as the human annotators would while explicitly capturing their annotation process in the learned model. 

How do we define mathematically that a post follows the \textit{yes} path or the \textit{no} path to a question $q_{i}$, i.e., we need to define exactly what $I_{yes}(q_{i})$ and $I_{no}(q_{i})$ means. Recalling our understanding of similarity between question and answer as being a proxy to answered as \textit{yes} or \textit{no}, we can use inner product based similarity between representations of question and post to determine $I_{yes}(q_{i})$ and $I_{no}(q_{i})$. For example, for a similarity model that takes as inputs representations for \textit{``Have you thought about being dead or what it is like to be dead''} and \textit{``Rarely is a day where I don't suffer from thoughts of self-harm''}, the output is a value indicative of high similarity relative to other input pairs. This is seen as the question \textit{``Have you thought about being dead or what it is like to be dead''} being answered as \textit{yes} by the response \textit{``Rarely is a day where I don't suffer from thoughts of self-harm''}. There are several options in Natural Language Processing (NLP) literature to construct representations of text
\begin{itemize}
    \item \textbf{Count Vectorizer}. Each sentence or text fragment is represented as counts of the words in the fragment padded with zeros according to largest fragment. Count vectorization however, does not consider the importance of words across different parts of the post, for example stop words might occur most frequently but provide little context.
    \item \textbf{TF-IDF}. TF-IDF corrects the defficiency of the Count Vectorizer method by adjusting counts by weighting for contextual importance across the post. However, TF-IDF still relies on exact matches of words being present or absent in the post.
    \item \textbf{Hashing Vectorizer}. Each sentence or fragement of the text is simply passed through a hash function. The idea is that similar fragments produce similar hash codes. The crytpic nature of the hash function (this is by design) is not amenable for interpretation or explainability analysis of the learned function.
    \item \textbf{Text Embeddings}. These are a set of neural network models that represent text in a vector space. Models such as word2vec, Transformer LMs such as GPT-3 and BERT are all examples of large neural networks that map the text to a vector space such that contextually similar texts are placed close together in the vector space while dissimilar texts are placed apart \cite{bert,gpt3,transformers}. Since, these are the state of the art and have shown remarkable effeciency and performance in recent years, we will use Transformer based LM representations for the text.
\end{itemize}

Note that the Text Embedding models provide vector representations of words. To construct a representation for the text fragment one might average the word representations contained in the fragment. However, this loses information about the order of the words and phrases in the text and hence we use a concatenation representation padded with zeros according to the longest text fragement. Thus generally we will denote a similarity function by $\mathcal{K}$ and representations of text $x$ and question $q_{i}$ using an embedding model as $x^{R}$ and $q_{i}^{R}$ respectively. Thus \[\Bigg(\pm\mathcal{K}\bigg(\frac{x_{sub}^{R}}{|x_{sub}^{R}|},\frac{q_{i}^{R}}{|q_{i}^{R}|}\bigg) \geq \pm\theta_{i}\Bigg)\] denotes the similarity between the text and question where $\theta_{i}$ are suitably chosen thresholds of accepted high similarity. The normalization of the representations by size is what makes an inner product a valid similarity measure in the range $-1~to~+1$. 
We will now formally develop the algorithm for \textbf{PK-iL}
\section{The PK-iL Algorithm}
We define a function that predicts the probability of post label being $Y=y$ according to the \textbf{PK} as follows:
\begin{equation}
\scriptsize
 P(Y=y\mid X=x) = \sum_{l\in Leaves}p_{l}\prod_{i=1}^{N_{q}}\lor_{x_{sub} \in x}\Bigg(\pm\mathcal{K}\bigg(\frac{x_{sub}^{R}}{|x_{sub}^{R}|},\frac{q_{i}^{R}}{|q_{i}^{R}|}\bigg) \geq \pm\theta_{i}\Bigg)   
\end{equation}
, where $p_{l}$ is defined as detailed in Section \ref{sec:treestruct}, $x_{sub}\in x$ is a fragment of the post $x$ (For example a sentence). $x_{sub}^{R}$, and $q_{i}^{R}$ are representations of $x_{sub}$ and $q_{i}$ from an embedding model, and $\mathcal{K}$ is a inner product function to measure similarity. 
\newline
\newline
$\pm$ signifies if we are checking if the question $q_{i}$ is answered as \textit{yes} or \textit{no} by fragment $x_{sub}$ in post $x$ with confidence $\theta_{i}$.
Using $\lor_{k=1}^{K}z_{k} = (\sum_{k=1}^{K}z_{k} \geq 0.5)$, we have:
\begin{equation}
\scriptsize
P(Y=y\mid X=x) = \sum_{l\in Leaves}p_{l}\prod_{i=1}^{N_{q}}\sum_{x_{sub} \in x}\Bigg(\pm\mathcal{K}\bigg(\frac{x_{sub}^{R}}{|x_{sub}^{R}|},\frac{q_{i}^{R}}{|q_{i}^{R}|}\bigg) \geq \pm\theta_{i}\Bigg) \geq 0.5
\label{eqn:PK-iL}
\end{equation}
We can then optimize the Bernoulli Loss $\mathcal{L}$ for an input post $X=x$ and label $Y=y$ is as follows:
\[\mathcal{L}(\{\theta_{i}\}_{i=1}^{N_{q}}) = P(Y=y|X=x)log(P(Y=y|X=x))\]
\[+\]
\[(1-P(Y=y|X=x))log(1-P(Y=y|X=x)) \]

We perform hyperparameter tuning to choose the embedding model, fragment size $x_{sub}$, and $\mathcal{K}$ (see Section {\ref{sec:exps}}). Since $\mathcal{L}(\{\theta_{i}\}_{i=1}^{N_{q}})$ is strongly convex, we use Newton's optimization method to learn the parameters of the model. The algorithm for \textbf{Process Knowledge infused Learning (PK-iL)} is as follows:
 \begin{algorithm}
\caption{Process Knowledge infused Learning (\textbf{PK-iL})}\label{alg:PKiL}
\begin{algorithmic}[1]
\State Compute $p_{l}~\forall$ leaves $l$ from the ground truth
\State Choose Kernel $\mathcal{K}$, fragment size, and \textbf{CE} model for representation
\State Initialize $\theta_{i},~\forall i \gets 1~to~N_{q}$
\For{$k \gets 1~to~K$} \Comment Begin Newton's method
\For{$\theta_{i}$, where $i \gets 1~to~N_{q}$}
\State Compute $\theta_{i}^{'} = \nabla_{\theta_{i}}\mathcal{L}(\theta_{i})$
\State Compute $\theta_{i}^{''} = \nabla\theta_{i}^{'} = \nabla_{\theta_{i}}(\nabla_{\theta_{i}}\mathcal{L}(\theta_{i}))$
\State Set $\theta_{i} = \theta_{i} - \frac{\theta_{i}^{'}}{\theta_{i}^{''}+1}$
\Comment add $1$ to avoid divide by zero error
\EndFor
\EndFor
\State return $\theta_{i},~\forall i \gets 1~to~N_{q}$
\end{algorithmic}
\end{algorithm}
Here we see that \textbf{PK-iL} is general enough to allow embedding models suitable to the task and \textbf{PK} suitable to any domain. However, in our experimental results we will evaluate \textbf{PK-iL} both quantitatively and qualitatively evaluation using the expanded \textbf{PK} enhanced CSSRS dataset (see Section \ref{sec:dataset}.

\paragraph{\textbf{Prediction:} } Prediction is carried out by choosing the summand in Equation \ref{eqn:PK-iL} that has the highest value once normalized by dividing by the sum of the summands, in order for it to be a probability.

\section{Experimental Setup and Results}{\label{sec:exps}}
For the LM to understand language in the context of suicidal thought patterns it needs to be fine-tuned on such a data. For this we word2vec representations on corpus of suicide related subreddits as as well as fine-tune LMs during training on the same corpus. Thus we obtain embeddings of the text contextualized to suicidal conversation in order to accurately infer \textit{yes} or \textit{no} from similarity. To implement the word2vec model, we use the gensim library and the Continuous Bag of Words (CBOW) model \cite{mikolov2013efficient}. Note that in the word2vec model due to lack of tokenization coverage as in LMs, we chunk the string one letter at a time and check against the list of words and their vectors. The LMs we fine-tune are:
\begin{itemize}
    \item XLNET - An auto-regressive language model in which the training objective calculates the probability of a token conditioned on all permutations of tokens in a fragment.  When trained on a very large data, the model achieves SOTA performance across several tasks in the GLUE benchmark \cite{yang2019xlnet}\cite{wang2018glue}. We use the default hyperparameters during training.
    \item RoBERTa - A BERT based model where the hyperparameters are further tuned for superior performance over BERT on GLUE benchmark tasks. We use the default hyperparameters during training RoBERTa \cite{liu2019roberta}. 
    \item Google T5 - A transformer model trained on various text-to-text tasks such as translation, summarization, etc. T5 demonstrate superior transfer capabilities across tasks in the GLUE benchmark. We use the T5-small model and the default parameters \cite{raffel2019exploring}.
    \item ERNIE - A transformer model trained on natural language corpora and large knowledge graphs and is thus a suitable model to consider in a structured knowledge intensive field like mental health. We use the default parameters for ERNIE \cite{zhang2019ernie}. 
    \item Longformer - A transformer model that excels at capturing long text inputs. As some of the posts can be over $8000$ characters long, the longformer is a suitable model to consider for our dataset. We use the default parameters for the long former \cite{beltagy2020longformer}. 
\end{itemize}
We believe the wide range of transformer architectures above are sufficient to test our approach. We train all our models on the Google Colab platform.
\paragraph{Inner Products} Bubeck et al., shown that $O(nd)$, where $n$ is the number of data points and $d$ is the true underlying data dimension \cite{bubeck2021universal}. The Transformer outputs are already high dimensional but Bubeck show that for natural language the models still need to get larger! Thus we use a popular trick to compute inner products in higher dimensions - the Kernel trick. Polynomial Kernel can project the data to very high dimensions and the Gaussian Kernel can project the data to an infinite number of dimensions. We see the use of a Kernel significantly improves the performance over simple cosine similarity (polynomial kernel of dimension $1$). In our experiments we use the Gaussian Kernel to compute the inner product.

For the fragment size we found a span of 1-2 sentences to be the best performing model for each transformer and kernel choice.  
\subsection{Quantitative Evaluation}
 For baseline accuracy we directly use the embedding models to predict the label as in a traditional machine learning pipeline. For word2vec, we use the representations of the post and pass it through a logistic regression model. We make a slight modification where weights for all entries for a single word vector are shared. Table \ref{table:quantitative} shows a comparison of accuracy for all the models with their baseline, \textbf{PK-iL} with Cosine Similarity, and \textbf{PK-iL} with a Gaussian Kernel. 
 
 \paragraph{Suicidality Context Capture}It is very interesting to note the word2vec, trained using the CBOW method, is the best performing model in the Baseline, Cosine Similarity, and the Gaussian Kernel case. We hypothesize upon inspection of the embeddings that word2vec, since trained from scratch on the suicide related post corpus captures contextual dependencies between suicidality tokens and phrases much better than LMs. LMs need to be fine-tuned on very large amounts of data to adapt against non suicidality term related contexts that they have trained on using massive corpora.

From our analysis we note that for domain specific tasks such as mental health related prediction, it is perhaps better to train contextual dependencies between words and phrases from scratch as pretrained models are already heavily biased towards the contextual dependencies on the corpora that they are trained on.

\paragraph{Comparing Baselines} Across all the models we see, \textbf{PK-iL} improves upon the accuracy of the baseline models by upto almost $15$\% points (for Longformer). Although to confirm our statement we have to rule out effects of collecting more data, adding/deleting features etc, using neural representations and limited data alone, explicitly controlling the learned model with process knowledge shows significant performance gains.
\paragraph{High Dimensional Data} Our experiments indeed show that even for domain specific corpora such as posts related to suicidality, the latent dimension of the text required to learn metric spaces is indeed very high. Improving the dimensionality shows little gain in this setting. But we hypothesize that for text from broader domains (e.g. text related to mental health in general), the dimensionality expansion will show more significant improvements.
\begin{table}[!h]
\footnotesize
\centering
\begin{tabular}{c|cccc} 
\toprule[1.5pt]
Model                                                                                      & Baseline 
& Cosine Similarity & Gaussian Kernel  \\ 
\midrule[1pt] \\
Word2Vec & \textbf{75}\% /~\textbf{69}\% & \textbf{83}\% /~\textbf{78}\% & \textbf{84}\% /~\textbf{72}\% \\
\hline \\
XLNET & 70\% / 65\% & 80\% / 69\% & \textbf{84}\% / 71\% \\
\hline \\
RoBERTa & 67\% / 62\% & 70\% / 64\% & 71\% / 62\% \\
\hline \\
T5 & 67\% / 54\% & 72\% / 52\% & 75\% / 64\% \\
\hline \\
ERNIE & 68\% / 62\% & 75\% / 69\% & 80\% / 71\% \\
\hline \\
Longformer & 50\% / 38\% & 65\% / 49\% & 67\% / 48\% \\
\bottomrule[1.5pt]
\end{tabular}
\caption{The mean accuracy/AUC-ROC, rounded up, of all the models - Column one shows the baseline where the model is directly used to predict the label, Column two shows \textbf{PK-iL} with Cosine Similarity for Kernel choice for each embedding model choice of representation, Column two shows \textbf{PK-iL} with a Gaussian Kernel for Kernel choice for each embedding model choice of representation}
\label{table:quantitative}
\end{table}
\subsection{Qualitative Evaluation} As mentioned earlier the qualitative evaluations among three expert annotators received a score of 0.7 agreement. Now, We will look at some of the explanations generated for interesting examples that show cases where \textbf{PK-iL} performed well and cases where it did not. We will also compare with explanations of the word2vec model which is easy to visualize using the weights of the word2vec vectors from the logistic regression model. We highlight the phrases whose individual word sums are greater than a threshold.
\paragraph{Post Example 1.} We will compare Word2vec baseline and \textbf{PK-iL} with the Gaussian Kernel.
\newline
\newline
\begin{table}[!ht]
    \centering
    \begin{tabular}{|p{7cm}|}
        \hline \\
        \textbf{Prediction:} Ideation \\
        \textbf{Ground Truth:} Indication \\
        \textbf{Model:} Word2Vec Baseline \\ 
        \hline \\
        'A book is usually what I do when Im \color{teal}getting down\color{black}, but it doesnt work when I start getting \color{teal}panicky\color{black}. Ill try the carbs, the caffeine doesnt work because Ive gotten it in a movie theater and had a soda with me...', 'A few reasons. I feel \color{teal}backed into a corner \color{black} mostly. And Im Tired of being Tired of everything. If that makes sense.', 'Thank you! I understand its a \color{teal}sad \color{black} thing. But I also want people to realize that there can be humor in anything and its the best way to deal with this. Its how I would do it. ', 'I really dont want to \color{teal}ask for help\color{black}. Id rather not let anyone know Im having these kind of issues.' \\
        \hline 
    \end{tabular}
    \caption{Example of attention visualization based explanations}
    \label{tab:my_label}
\end{table}

From this example, we can clearly see word2vec associating phrases and words that characterize a low mood with suicidal ideation. In real life such words may raise triggers in the minds of a clinician and may benefit their analysis. However, the human annotator seems to have labeled this as indication based on the ``there can be humor in everything'' part of the post. 
\begin{table}[!ht]
    \centering
    \begin{tabular}{|p{7cm}|}
        \hline \\
        \textbf{Prediction:} Indication \\
\textbf{Ground Truth:} Indication \\
\textbf{Model:} \textbf{PK-iL} with Gaussian Kernel \\ 
        \hline \\
         'A book is usually what I do when Im \color{black}getting down\color{black}, but it doesnt work when I start getting \color{black}panicky\color{black}. Ill try the carbs, the caffeine doesnt work because Ive gotten it in a movie theater and had a soda with me...', 'A few reasons. I feel \color{black}backed into a corner mostly. And Im Tired of being Tired of everything. If that makes sense.', 'Thank you! I understand its a \color{black}sad \color{black} thing. \color{teal}But I also want people to realize that there can be humor in anything and its the best way to deal with this\color{black}. Its how I would do it. ', 'I really dont want to \color{black}ask for help\color{black}. Id rather not let anyone know Im having these kind of issues.' \\
        \hline \\
        \textbf{Explanation:} 1. Wish to be dead (\textit{no}) $\rightarrow$ indication \\
        \hline
    \end{tabular}
    \caption{Example of explanation based on PK-iL}
    \label{tab:my_label}
\end{table}
Recall that \textbf{PK-iL} deals with whole fragements of text and can therefore never highlight phrases as we experimented with fragment lenghts of \mbox{1-3} sentences. The highest threshold among the similarity functions in Equation \ref{eqn:PK-iL} corresponded to the fragment highlighted and the path 1. Wish to be dead \textit{no} and hence the model picks indication with probability equal to inter-annotator agreement of $y=indication$ at that leaf. Such an explanation although subject to annotator agreements is more informative to the clinician about the models prediction. Although, the highlights from the Word2vec model provide important cues as to the user's suicidal thought patterns it is unclear to the clinician why certain words were highlighted and certain others ignored. For example, why just ``panicky'' and not the whole phrase ``getting panicky''?. 

\paragraph{Embeddings vs PK-iL explanations - Developer vs End-user Perspective:}
Computer scientists with deep understanding of logistic regression weights and biases may find the embedding model based visualization easier to understand and replicate. They would clear understand the contextual dependencies between tokens and phrases learned by the inner mechanism of Word2vec and could therefore reasonably expect if the weights of a token or phrase are high in logistic regression, contextually related words will also be high. Also, that roughly statistically frequent tokens and the most frequent co-occurring words, per class label, are most likely to be highlighted in the model explanation. This makes perfect sense to the developer. However, the domain expert will struggle to palate the idea of statistically likely words and frequently co-occurring words as a valid explanation for the prediction.

\paragraph{Post Example 2.} We will see \textbf{PK-iL} with Gaussian Kernel outputs for a slightly more interesting example.
\begin{table}[!ht]
    \centering
    \begin{tabular}{|p{7cm}|}
        \hline \\
        \textbf{Prediction:} Behavior or Attempt \\
\textbf{Ground Truth:} Behavior or Attempt \\
\textbf{Model:} \textbf{PK-iL} with Gaussian Kernel \\
        \hline \\
         'I wish I could give a shit about what would make it to the front page. \color{teal}I have been there and got nothing. Same as my life. I do have a gun.\color{black}', 'I thought I was talking about it. \color{teal}I am not on a ledge or something, but I do have my gun in my lap.', \color{black} 'No. I made sure she got an education and she knows how to get a job. I also have recently bought her clothes to make her more attractive. She has told me she only loves me because I buy her things. '  \\
        \hline \\
        \textbf{Explanation:} 1. Wish to be dead (\textit{yes}) $\rightarrow$ 2. Non-Specific Active Suicidal Thoughts (\textit{yes}) $\rightarrow$ Active Suicidal Ideation with Some Intent to Act, without Specific Plan (\textit{yes}) $\rightarrow$ Behavior or Attempt \\
        \hline
    \end{tabular}
    \caption{Example of explanation based on PK-iL for example 2}
    \label{tab:my_label}
\end{table}

Here we can see how \textbf{PK-iL} highlights multiple sentences that satisfy its explanation generated. Note that the Word2Vec models prediction was also correct in this instance highlighting phrases such as ``On a ledge'', ``have a gun'' and ``gun in my lap''.
\paragraph{Correctness of Prediction} For the example post considered, the correctness of the prediction is subject to interpretation by human experts. This is why there is interannotator disagreement. \textbf{PK-iL} is however theoretically capable of performing as well as the annotators as per Assertion \ref{rem:first} in the best case. The intutions behind \textbf{PK-iL} focus more on understanding the experts thought process and providing explanations that they can understand rather than on the correctness of prediction.

Thus we believe fundamental algorithmic and data annotation changes like the \textbf{PK-iL} paradigm will result in faster integration of assistive machine learning technology in real-world applications. 

\section{Conclusion and Future Work}
In this study we develop a novel paradigm \textbf{PK-iL} that introduces the need for richer annotation and high performance explicit process guided explanation models that the end-user can readily understand. The dataset contains a lot of noisy and long posts. In such settings both \textbf{PK-iL} and embedding models performed poorly. These inherent challenges of social media data will need to be addressed in future work. Additionally, \textbf{PK-iL} also has the potential to identify regions of the example space that the \textbf{PK} applies to with high inter-annotator agreement. This can assist in soliciting more refined guidelines on those cases where scales such as the CSSRS clearly do not work. While these scales have been developed over decades of research, machine learning techniques such as \textbf{PK-iL} have the potential to provide assistive refinement of existing and established guidelines. 

\section{Acknowledgements}
This research is support by National Science Foundation (NSF) Award \# 2133842 “EAGER: Advancing Neuro-symbolic AI with Deep Knowledge-infused Learning,” . Any opinions, findings, and conclusions or recommendations expressed in this material are those of the author(s) and do not necessarily reflect the views of the NSF.

\bibliography{ijcai22}
\bibliographystyle{named}
\end{document}